\documentclass{article}

\usepackage{microtype}
\usepackage{graphicx} \usepackage{booktabs}
\usepackage{microtype}
\usepackage{amsmath}
\usepackage{amssymb}
\usepackage{verbatim}
\usepackage{float}

\usepackage{hyperref}
\usepackage{caption}
\usepackage{subcaption}

\newcommand{\mc}[1]{\mathcal{#1}}

\usepackage[accepted]{icml2018}

\icmltitlerunning{Progress \& Compress: A scalable framework for continual learning}

\begin{document}

\twocolumn[
\icmltitle{Progress \& Compress: A scalable framework for continual learning}

\icmlsetsymbol{equal}{*}

\begin{icmlauthorlist}
\icmlauthor{Jonathan Schwarz}{dm}
\icmlauthor{Jelena Luketina}{ox}
\icmlauthor{Wojciech M. Czarnecki}{dm}
\icmlauthor{Agnieszka Grabska-Barwinska}{dm}
\icmlauthor{Yee Whye Teh}{dm}
\icmlauthor{Razvan Pascanu}{equal,dm}
\icmlauthor{Raia Hadsell}{equal,dm}
\end{icmlauthorlist}

\icmlaffiliation{dm}{DeepMind, London, United Kingdom}
\icmlaffiliation{ox}{Department of Computer Science, University of Oxford, Oxford, United Kingdom}

\icmlcorrespondingauthor{Jonathan Schwarz}{schwarzjn@google.com}
\icmlcorrespondingauthor{Razvan Pascanu}{razp@google.com}

\icmlkeywords{Continual Learning, Transfer Learning, Deep Learning}

\vskip 0.3in
]

\printAffiliationsAndNotice{\icmlEqualContribution}  

\def\KB{\mathsf{KB}}

\begin{abstract}
We introduce a conceptually simple and scalable framework for continual learning domains where tasks are learned sequentially. Our method is constant in the number of parameters and is designed to preserve performance on previously encountered tasks while accelerating learning progress on subsequent problems. This is achieved by training a network with two components: A \textit{knowledge base}, capable of solving previously encountered problems, which is connected to an \textit{active column} that is employed to efficiently learn the current task. After learning a new task, the active column is distilled into the knowledge base, taking care to protect any previously acquired skills. This cycle of active learning (progression) followed by consolidation (compression) requires no architecture growth, no access to or storing of previous data or tasks, and no task-specific parameters. We demonstrate the progress \& compress approach on sequential classification of handwritten alphabets as well as two reinforcement learning domains: Atari games and 3D maze navigation.
\end{abstract}

\section{Introduction}
\label{sec:intro}

The standard learning process of neural networks is underpinned by the assumption that training examples are drawn i.i.d.\ from some fixed distribution. In many scenarios such a restriction is not of major concern. However, it can prove to be an important limitation particularly when a system needs to continuously adapt to a changing environment, as often happens in reinforcement learning and other interactive domains such as robotics or dialogue systems. This ability to learn consecutive tasks without forgetting how to perform previously trained problems is known as continual learning \citep[e.g.][]{Ring1995}.

A large body of literature recognises the importance of the continual learning problem, and there has been some increased interest in the topic recently  \citep[e.g.][]{rusu2016progressive, shin2017continual, lopez2017gradient, nguyen2017variational, kirkpatrick2017overcoming, chaudhry2018riemannian}. Part of the challenge stems from the fact that there are multiple, often competing, desiderata for continual learning:

(i)~A continual learning method should not suffer from \emph{catastrophic forgetting}. That is, it should be able to perform reasonably well on previously learnt tasks. 
(ii)~It should be able to learn new tasks while taking advantage of knowledge extracted from previous tasks, thus exhibiting \emph{positive forward transfer} to achieve faster learning and/or better final performance. (iii)  It should be \emph{scalable}, that is, the method should be trainable on a large number of tasks. (iv) It should enable \emph{positive backwards transfer} as well, which means gaining immediate improved performance on previous tasks after learning a new task which is similar or relevant. (v) Finally, it should be able to learn without requiring task labels, and ideally it should even be applicable in the absence of clear task boundaries.

Many approaches address some of these to the detriment of others. For example: Naive finetuning often leads to successful positive transfer, but suffers from catastrophic forgetting; 
elastic weight consolidation (EWC) \cite{kirkpatrick2017overcoming} focuses on overcoming catastrophic forgetting but the accumulation of Fisher regularisers can over-constrain the network parameters leading to impaired learning of new tasks;
Progressive Networks \cite{rusu2016progressive} avoid catastrophic forgetting altogether by construction, however it suffers from lack of scalability  as the network size scales quadratically in the number of tasks.

\begin{figure}[!ht]
\centering
\includegraphics[width=1.0\columnwidth]{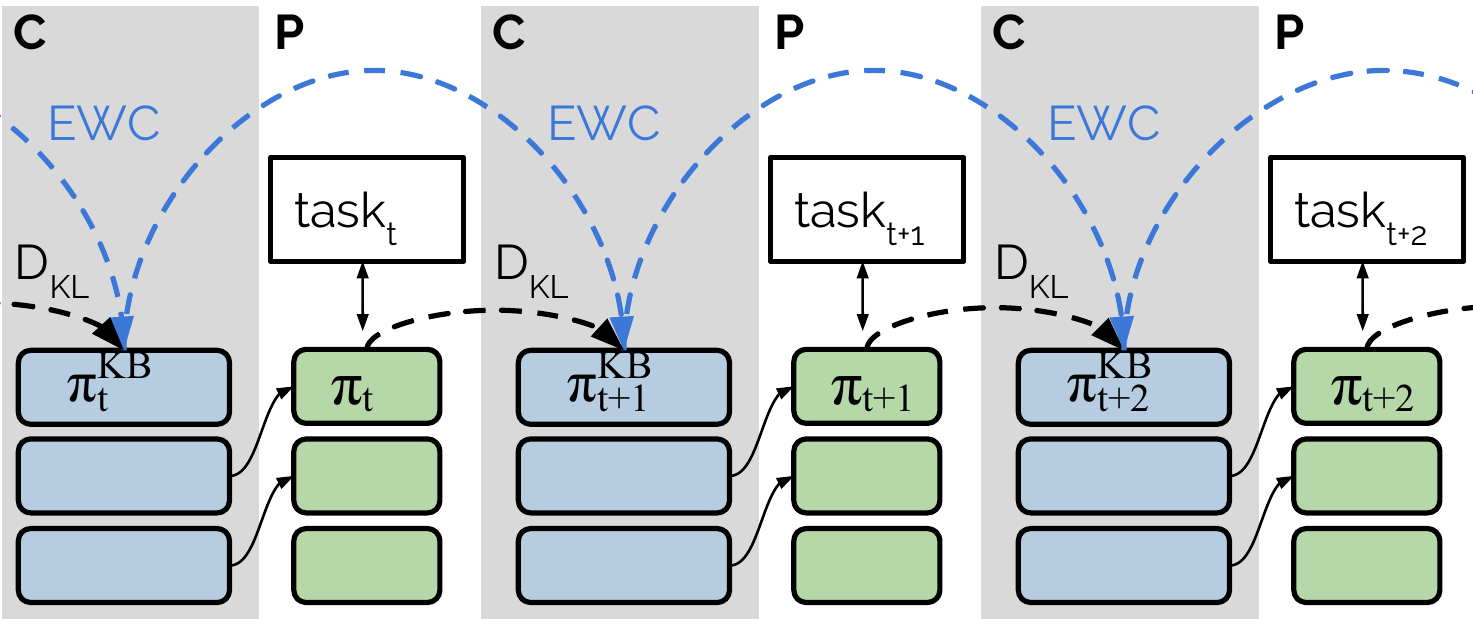}
\caption{Illustration of the Progress \& Compress learning process. In the compress phases (C), the policy learnt most recently by the active column (green) is distilled to the knowledge base (blue) while protecting previous contents with EWC (Elastic Weight Consolidation). In the progress phases (P), new tasks are learnt by the active column while reusing features from the knowledge base via lateral, layerwise connections.}
\label{fig:p_and_c_overview}
\vspace{-1em}
\end{figure}

This paper presents a step towards unifying these techniques in a framework that satisfies multiple desiderata, by taking advantage 
of their complementary strengths while minimising their weaknesses.
The proposed method implements two neural networks, a \textit{knowledge base} and an \textit{active column}, which are trained in two distinct, alternating phases. During the \emph{progress} phase, the network is presented with a new learning problem, and only parameters in the active column are optimised. Similar to the architecture of Progressive Networks \cite{rusu2016progressive}, layerwise connections between the knowledge base and the active column are added to facilitate the reuse of features encoded in the knowledge base, thus enabling positive transfer from previously learnt tasks.  
At the completion of the progress phase, the active column is distilled into the knowledge base, thus forming the \textit{compress} phase. During distillation, the knowledge base must be protected against catastrophic forgetting such that all previously learnt skills are maintained. We propose a modified version of Elastic Weight Consolidation  \cite{kirkpatrick2017overcoming} to mitigate forgetting in the knowledge base. 
The Progress \& Compress (P\&C) algorithm alternates these two phases, allowing new tasks to be encountered, actively learned, and then carefully committed to memory. The approach is purposefully reminiscent of daytime and nighttime, and of the role that sleep plays in memory consolidation in humans, allowing the important skills mastered during the day to be filed away at night. As P\&C uses two columns of fixed sizes, it is scalable to a large number of tasks. In experiments, we observe positive transfer, while minimising forgetting, on a variety of domains.

\section{The Progress and Compress Framework}
\label{sec:p_and_c}

The P\&C architecture is composed of two components, a \emph{knowledge base} and \emph{active column}. Both components can be visualised as columns of network layers which compute either predicted class probabilities (in case of supervised learning) or policies/values (in case of reinforcement learning). The two components are learnt in alternating phases (progress/daytime and compress/nighttime).  Figure \ref{fig:p_and_c_overview} provides an illustration of the architecture and the two phases of learning when P\&C is applied to reinforcement learning. 

\subsection{Learning a new task}

The separation of the architecture into two components allows P\&C to focus on positive transfer when a new task is introduced. As illustrated in Figure \ref{fig:p_and_c_overview}, the knowledge base (light blue) is fixed, while parameters in the active column (green) are optimised without constraints or regularisation, allowing effective learning on the new task. In addition, P\&C enables the reuse of past information through simple layerwise adaptors to the knowledge base (lateral arrows), an idea borrowed from Progressive Nets. 

Adaptors themselves are implemented as multi-layer perceptrons. Specifically, if $h_i$ denotes the activations in layer $i$, superscript~$^\KB$ the knowledge base, and $\sigma$ a nonlinearity, the $i$th layer of the active column is computed as:
\begin{align}
  h_i = \sigma(W_ih_{i-1}+\alpha_{i} \odot U_{i}\sigma(V_i h_{i-1}^\KB + c_i) + b_i)
  \label{eq:adapter}
\end{align}
where $b_i$ and $c_i$ are biases, $\alpha_i$ is a trainable vector of size equal to the number of units in layer $i$, $W_i, U_i, V_i$ are weight matrices and $\odot$ denotes elementwise multiplication. The vector $\alpha_i$ is initalised by sampling from $\mc{U}(0,0.1)$.
In the case of convolutional networks, we use $1\times1$ convolutions for the adaptors. 

Note that one could make this phase similar to naive finetuning of a network trained on previous tasks by not resetting the active column or adaptors upon the introduction of a new task. Empirically, we found that this can improve positive transfer when tasks are very similar. For more diverse tasks however, we recommend re-initialising these parameters, which can make learning more successful.

\subsection{Distillation and knowledge preservation}
\label{sec:distillation}

During the ``compress" phase, newly learnt behaviour is consolidated into the knowledge base. This is also where methods guarding against catastrophic forgetting are introduced.
The consolidation is done via a distillation process \cite{hinton2015distilling, rusu2015policy}, which is an effective mechanism for transferring knowledge from the active column to the knowledge base. In the RL setting it has the additional advantage that the scale of the distillation loss does not depend on the (scale of the) reward scheme, which can be quite different for different tasks. We minimise the cross-entropy between the teacher's (active column) and student's (knowledge base) prediction/policy.

As a method of choice for knowledge preservation, we rely on Elastic Weight Consolidation (EWC) \cite{kirkpatrick2017overcoming}, a recently introduced method that poses an approximate Bayesian solution to continual learning. The main insight is that information pertaining to different tasks can be incorporated sequentially into the posterior without suffering catastrophic forgetting since the resulting posterior does not depend on task ordering. However, the exact posterior is intractable for neural networks, and EWC employs a tractable Gaussian approximation.  This results in regularisation terms, one for each previous task, that constrain the parameters not to deviate too much from those that were optimised. However, the number of regularisation terms grow linearly in the number of tasks, meaning that the original EWC algorithm is not scalable to a large number of tasks.  In Section \ref{sec:EWC}, we elaborate on a modification that we refer to as \emph{online EWC} which does not exhibit this linear growth in computational requirements. 

In summary, for consolidating task $k$ into the knowledge base, we optimise the following loss with respect to the parameters $\theta^\KB$ of the knowledge base while keeping the parameters of the active column unchanged,
\begin{align}
\mathbb{E}\Big[\mathsf{KL}(\pi_k(\cdot|x)\|\pi^\KB(\cdot|x))\Big] + \frac{1}{2} \|\theta^\KB-\theta^\KB_{k-1}\|^2_{ \gamma F^*_{k-1}}
\end{align}
where $\pi_k(\cdot|x)$ and $\pi^\KB(\cdot|x)$ are the prediction/policy of the active column (after learning on task $k$) and knowledge base respectively, $x$ is the input, $\mathbb{E}$ denotes expectation over either the dataset or the states of the environment under the active column, $\theta^\KB_{k-1}$ and $F^*_{k-1}$ are the mean and diagonal Fisher of the online EWC Gaussian approximation resulting from previous tasks, and $\gamma$ is a hyperparameter (see Section \ref{sec:EWC}). The policies are computed at inverse temperature $\tau$ (a hyperparameter). Note that  $\pi_k$ is fixed throughout the consolidation process to that learnt on task $k$.

\section{Related Work}
\label{sec:related_work}

We now provide a brief survey of work in the areas of continual learning, characterising each approach in the light of the desiderata introduced in Section \ref{sec:intro}. Note that  continual learning is known by different names (though with somewhat different foci), such as lifelong learning \cite{thrun1996explanation} and never-ending learning. Slightly different than continual learning, different aspects of \emph{transfer} learning for reinforcement learning are discussed and compared in \citet{AAAIMag11-Taylor}.

A common method of choice is finetuning a pretrained model on a target domain, hence introducing an alternative method of initialisation. This is commonly used due to its simplicity and has been shown to be a successful method for positive transfer, provided there is sufficient task similarity. Early successful applications include unsupervised to supervised transfer learning \cite{bengio2012deep} and various results in the vision domain. When a sequence of tasks is considered, this is usually done through the careful design of curricula, introducing tasks of increasing complexity. As catastrophic forgetting is a significant issue, such methods are usually not able to compose skills learned in previous tasks unless such skills keep being reused. Examples of this technique include transfer from Deep Q-Networks \cite{parisotto2015actor} or curriculum learning in memory models \cite{graves2016hybrid}.

A second family of methods introduces task-specific parameters, allowing components within a larger ensemble to learn representations of the data specific to a given task. Transfer in such models can be achieved by sharing a subset of features or by introducing connections between such task-specific modules. An apparent issue with these methods is their lack of scalability, often making the application to large number of tasks computationally cumbersome and unstable. In addition, a task label has to be either provided or inferred at test time such that the correct module can be chosen.

Progressive networks \cite{rusu2016progressive} are a method within this category designed for continual learning. The authors propose an architecture that introduces an identical "neural network column" for each task, allowing transfer through adaptor connections to columns dedicated to previous problems. The method has particular appeal, namely its immunity against catastrophic forgetting, which is due to freezing parameters after a task has been learnt. Unfortunately, this does not allow for positive backward transfer. 

Learning Without Forgetting \cite{li2017learning} mainly focuses on improving resilience against catastrophic forgetting. This is achieved by recording the output of \textit{old} task modules on data from the \textit{current} task before any update to the shared parameters, allowing regularisation towards those values during training. A problem with this method is that it is not immediately applicable to Reinforcement Learning. Other examples include \cite{aljundi2016expert} which introduces a gating mechanism between columns and \cite{rozantsev2016beyond}, who formulate an alternative regularisation objective to keep weights of columns tied. 

Another category of work is based on the idea of episodic memory, where examples from prior tasks are stored to effectively recall experience encountered in the past \cite{robins1995catastrophic}. Examples making use of this idea are \cite{rebuffi2016icarl, schmidhuber2013powerplay, thrun1996explanation}. A similar approach is proposed by
\citet{lopez2017gradient}, however instead of storing examples, gradients of the previous task are stored, such that at any point in time the gradients of all tasks except the current one can be used to form a trust region that prevents forgetting. Such methods can be effective against catastrophic forgetting, provided a good mechanism for the selection of relevant experience is proposed. An inherent problem is the constraint on the amount of experience that can be stored in memory, which could quickly become a limiting factor in large scale problems. Recent methods have tried to overcome this by sampling synthetic data from a generative model \cite{shin2017continual, silver2013lifelong}. This shifts the catastrophic forgetting problem to the training of the generative model.

The replay of past experience can be seen as moving closer to \emph{multitask} learning \cite{caruana1998multitask}, which differs from continual learning in that data from all tasks is available and used jointly for training. In the simplest case, this is achieved by sharing parameters, similar to aforementioned methods. Distral \cite{teh2017distral} explicitly focuses on positive transfer through sharing a \emph{distilled} policy which captures and transfers behaviour across several tasks. 
Distillation is also used by \citet{divideconquerrl2017} to composite multiple low-level RL skills, and by \citet{FurlanelloZSIT16} to maintain performance on multiple sequential supervised tasks through a transfer learning paradigm. 

Another family of methods avoid catastrophic forgetting by regularising learning. One prominent example in this category is Elastic Weight Consolidation \cite{kirkpatrick2017overcoming}. Synaptic Intelligence \cite{zenke2017continual} is similar to Elastic Weight Consolidation but computes an importance measure online along an entire learning trajectory. Recently \citet{he2018} proposed a different mechanism, which employs a projection of the gradients such that no direction relevant to the previous task is affected. 

PLAiD \cite{berseth2018} is method with similarities with our approach. However the method is not designed for continual learning, but rather for maximising transfer, since it assumes access to all tasks at any point in time. The approach relies on two stages, similar to ours. In one stage a new task is learnt, transferring from the previous learnt tasks. In the second stage, the learnt policy is consolidated by multitask distillation from all previously seen tasks. 

Some ideas that could serve as inspiration for future work on the continual learning problem can also be found in \citet{schmidhuber2018one}.

\section{Online EWC}
\label{sec:EWC}

\def\task{\mathcal{T}}

The starting point of Elastic Weight Consolidation (EWC) \cite{kirkpatrick2017overcoming} is an approximate Bayesian treatment of continual learning. Let $\theta$ be the parameter vector of interest (in P\&C these are the parameters $\theta^\KB$ of the knowledge base; i.e.\ we drop the superscript $^\KB$ in this section for simplicity), and let $\task_{1:k} = (\task_1,\task_2,\ldots,\task_k)$ denote the data associated with a sequence of $k$ tasks. The posterior of $\theta$ is:
\begin{align}
p(\theta|\task_{1:k}) &\propto
p(\theta)\prod_{i=1}^k p(\task_i|\theta)\label{eq:ewclikelihood}\\
&\propto p(\theta|\task_{1:k-1})p(\task_k|\theta)
\label{eq:ewcposterior}
\end{align}
where the multi-task likelihood term in Eq. (\ref{eq:ewclikelihood}) factorises due to the task data conditional independence.
According to Eq. \eqref{eq:ewcposterior}, the posterior given all $k$ tasks can be computed sequentially, by first computing that for the first $k-1$ tasks, and treating it as the conditional prior as we incorporate the likelihood for the $k$-th task.

Unfortunately, the exact posteriors needed are intractable, and are replaced by Laplace's approximation \cite{mackay2003information}. For EWC: 
\begin{align}
p(\task_i|\theta) \approx \mathcal{N}(\theta; \theta^*_i,  F^{-1}_i),
\end{align}
with mean $\theta^*_i$ centred at the maximum a posteriori (MAP) parameter when learning task $i$, and precision given by the (diagonal) Fisher information matrix evaluated at $\theta^*_i$, which is a surrogate for the Hessian of the negative log likelihood that is guaranteed to be positive semidefinite. 

The MAP parameter is computed using a standard stochastic optimiser applied to the loss
\begin{align}
-\log p(\task_i|\theta) + \frac{1}{2}
\sum_{j=0}^{i-1}  \|\theta - \theta^*_j\|^2_{F_j}
\label{eq:ewcloss}
\end{align}
which is the negative log of~\eqref{eq:ewcposterior}.
The $j=0$ term is a notational convenience for the prior $-\log p(\theta)$ while the norm is the Mahalanobis norm. 

Note that in the above formulation a mean and a Fisher need to be kept for each task, which makes the computational cost linear in the number of tasks. One can reduce the cost to a constant by ``completing the square'' for the Fisher regularisation terms in \eqref{eq:ewcloss}.  Alternatively,
as pointed out by \cite{huszar2017quadratic},
we can apply Laplace's approximation to the whole posterior  \eqref{eq:ewcposterior}, rather than the likelihood terms.  
This results in the following loss:
\begin{align}
-\log p(\task_i|\theta) + \frac{1}{2}
 \|\theta - \theta^*_{i-1}\|^2_{\sum_{j=0}^{i-1} F_j}
\label{eq:ewcloss2}
\end{align}
Compared with \eqref{eq:ewcloss}, the difference is that the Gaussian approximation of previous task likelihoods are ``re-centred'' at the latest MAP parameter $\theta^*_{i-1}$. This means that we only need to keep the latest MAP parameter along with a running sum of the Fishers (which is another approximation, as Fisher information is a local measure and all $F_i$'s should more correctly be recomputed for the new $\theta^*$, an infeasible computational burden).  

Note that it is unclear what the effect of this re-centring will be for nonlinear neural networks (\cite{huszar2017quadratic} did not show any experimental validation). Shifting the mean to the latest MAP value will mean that older tasks will be remembered less well, since there will not be any regularisation terms constraining the parameters to be close to that learnt on the older tasks. We demonstrate this effect in the Appendix.

In a continual or life-long learning regime, where the model is applied to many tasks, one interesting aspect occurs when tasks can be revisited.  \cite{huszar2017quadratic} propose taking the expectation propagation (EP) \cite{minka2001expectation} approach of keeping an explicit approximation term for each likelihood, so that when a task is revisited the approximation to its likelihood can be removed and recomputed. However this means a return to the linear scaling of the original EWC. We will instead take a stochastic EP \cite{li2015stochastic} approach, which does not keep explicit approximations for each factor. Instead a single overall approximation term is maintained and updated partially when a task is revisited. More precisely, let $\theta^*_{i-1}$, $F^*_{i-1}$ be the MAP parameter and overall Fisher after presentation of $i-1$ tasks. The loss for the $i$th task is then:
\begin{align}
    -\log p(\task_i|\theta) + \frac{1}{2}\| \theta-\theta^*_{i-1}\|^2_{\gamma F^*_{i-1}}
    \label{eq:ewc_loss}
\end{align}
where $\gamma<1$ is a hyperparameter associated with removing the approximation term associated with the previous presentation of task $i$. If $\theta^*_i$ is the optimised MAP parameter and $F_i$ the Fisher for task $i$, the overall Fisher is then updated as 
\begin{align}
F^*_i = \gamma F^*_{i-1} + F_i\label{eq:ewc_acc_fish}
\end{align}
This approach has the benefit of avoiding the need for identifying the task labels, since the method treats all tasks equivalently (as opposed to EWC/EP). Identifying task boundaries is significantly easier than identifying task ids, since detection of a change in low level statistics is often sufficient. In the case of reinforcement learning, for example, changes in reward statistics can be used, which intuitively has connections to memory consolidation in the brain due to changes in dopamine levels.  Another interesting side effect is that the method can, via the $\gamma$ down-weighting, explicitly forget older tasks in a graceful and controlled (rather than catastrophic) manner.  This is useful, for example, if the learning has not converged on an older task, and it is better to gracefully forget its effect on the approximation term.  Graceful forgetting is also an important component for continual learning as forgetting older tasks is necessary to make space for learning newer ones, since our model capacity is fixed. Without forgetting, EWC misbehaves when the model runs out of capacity, as discussed in \cite{kirkpatrick2017overcoming}.  We refer to our modified method as \emph{online} EWC.

Finally one important observation we make is that each EWC penalty protects the policy in expectation over the state space, regardless of the reward scheme of the task. 
One problem that we can address is that it favours policies that are more deterministic, as in expectation, small changes to $\theta$ for such policies will cause larger changes in the KL and the Fisher matrix measures the curvature of the KL term.
This results in Fisher matrices of variable norm. However, the goal of the algorithm is to protect each task equally.

We counteract this issue by normalising the Fisher information matrices $F_i$ for each task. This allows the algorithm to compute the updates $F^*$ (Eq. \ref{eq:ewc_acc_fish}) based on the relative importance of weights in a network, i.e. treating each task equally rather than through an arbitrary scale of the original Fisher matrix.

\section{Experiments and Results}

We now provide an assessment of the suitability of P\&C as a continual learning method, conducting experiments to test against the desiderata introduced in Section \ref{sec:intro}. We introduce experiments varying in the nature of the learning task, their difficulty and the similarity between tasks. To evaluate P\&C for supervised learning, we first consider the sequential learning of handwritten characters of 50 alphabets taken from the Omniglot dataset \cite{lake2015human}. Considering each alphabet as a separate task, this gives us a way to test continual learning algorithms for their scalability.\footnote{Note that reported performance is not directly comparable to published results as we do not learn Omniglot in a few-shot learning setup.}

Assessing P\&C under more challenging conditions, we also consider the sequential learning of 6 games in the Atari suite \cite{bellemare2012investigating} (``Space Invaders'', ``Krull'', ``Beamrider'', ``Hero'', ``Stargunner'' and ``Ms. Pac-man''). This is a significantly more demanding problem, both due to the high task diversity and the generally more difficult RL domain. Specifically, the high task diversity constitutes a particular challenge for methods guarding against catastrophic forgetting. 

We also evaluate our method on 8 navigation tasks in 3D environments inspired by experiments with Distral \cite{teh2017distral}. In particular, we consider mazes where an agent experiences reward by reaching a goal location (randomised for each episode) and by collecting randomly placed objects along the way. We generate 8 different tasks by varying the maze layout, thus providing environments with significant task similarity. As the experiments with Distral show high transfer between these tasks, this allows us to test our method for forward transfer.

We use a distributed variant of the actor-critic architecture \cite{sutton1998reinforcement} for both RL experiments. Specifically, we learn both policy $\pi(a_t|s_t;\theta)$ and value function $V(s_t;\phi)$ from raw pixels, with $\pi, V$ sharing a convolutional encoder. All RL results are obtained by running an identical experiment with 4 random seeds. Training and architecture details are given in the Appendix.
For the remainder of the section, when writing P\&C, we assume the application of online EWC on the knowledge base. As a simple baseline, we provide results obtained by learning on a new task without protection against catastrophic forgetting (terming this ``Finetuning''\footnote{Referred to as ``SGD'' in \cite{kirkpatrick2017overcoming}}). Confidence intervals (68\%) appearing in several results throughout this section are calculated with a non-parametric bootstrap unless otherwise stated.

Throughout this section, we aim to answer the following questions: (i) To what extent is the method affected by catastrophic  forgetting? (ii) Does P\&C achieve positive transfer? (iii) How well does the knowledge base perform across all tasks after learning?

\subsection{Resilience against catastrophic forgetting}

\begin{figure}
    \centering
    \begin{subfigure}[b]{0.45\textwidth}
        \centering
        \includegraphics[width=\linewidth]{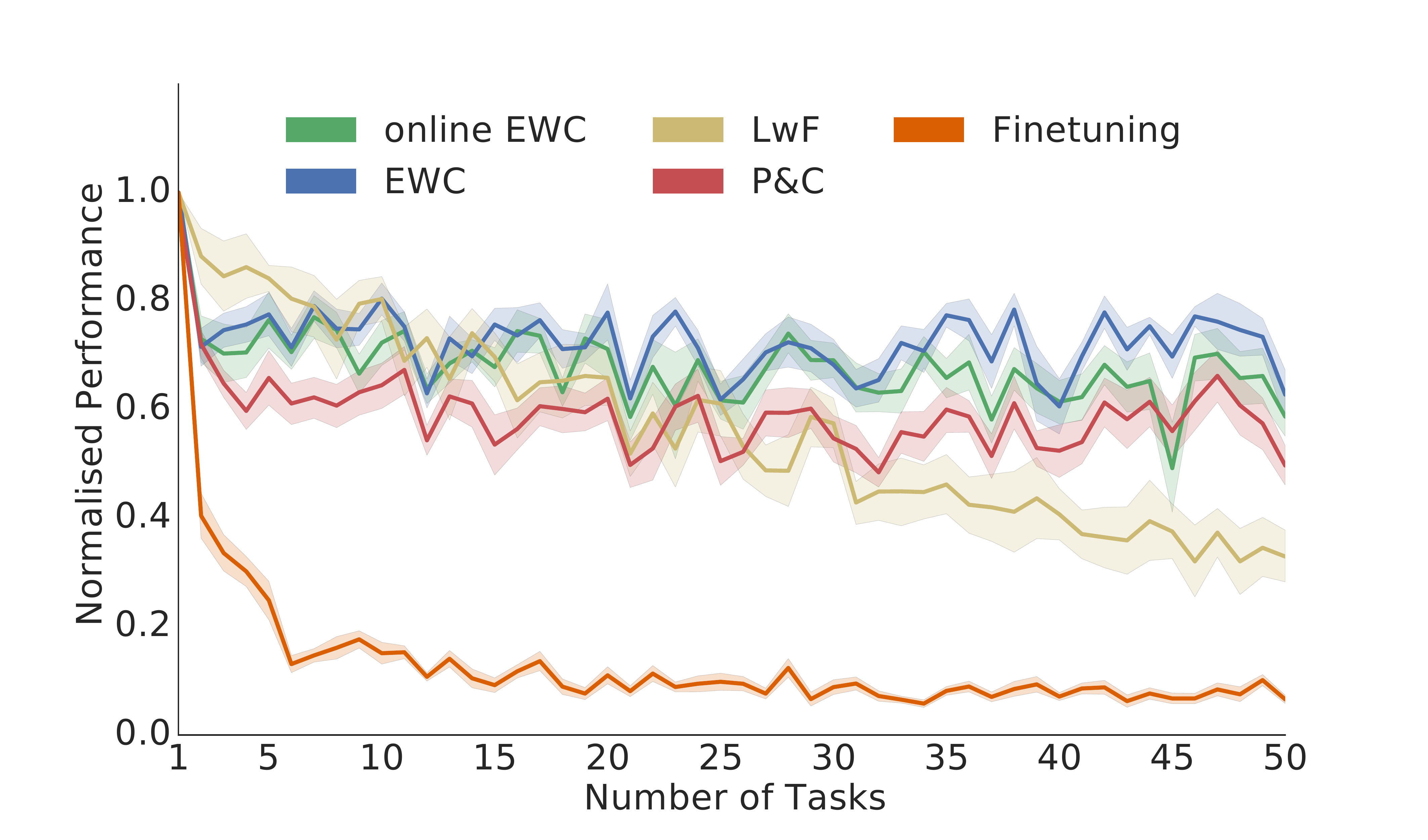}
        \caption{Performance retention: Results show how the accuracy
                 on an initial task changes as further alphabets are 
                 being learnt. Averaged over 5 different initial tasks.}
        \label{fig:omniglot_forgetting}
    \end{subfigure}
    \hfill
    \begin{subfigure}[b]{0.45\textwidth}
        \centering
        \includegraphics[width=\linewidth]{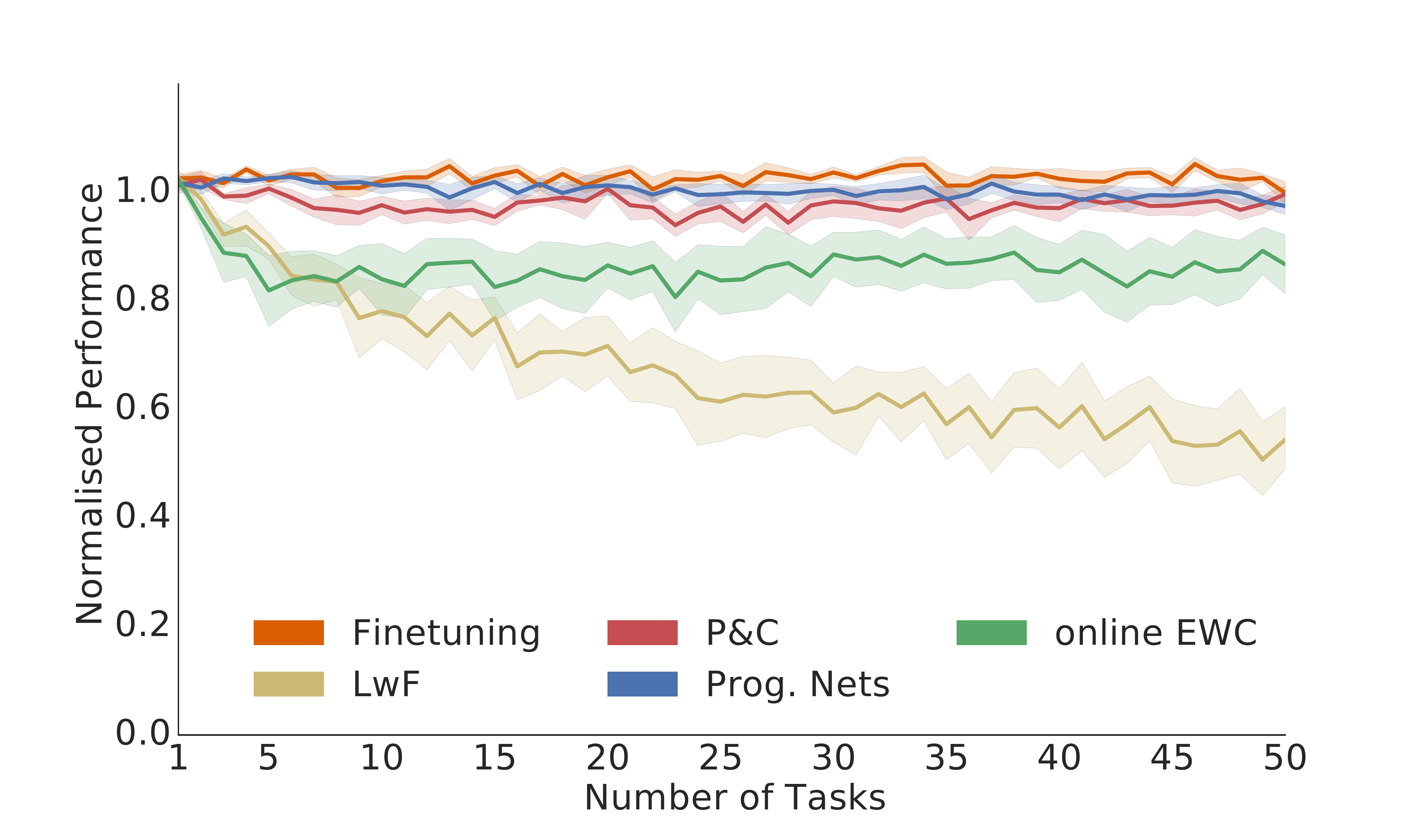}
        \caption{Forward transfer: Results show the relative performance achieved
                 on a unique tasks after a varying number of previous tasks have been
                 learnt. Averaged over 5 different final tasks.}
        \label{fig:omniglot_forward_transfer}
    \end{subfigure}
    \caption{Results on Omniglot. Performance normalised by training a single model on each task. Best viewed in colour.}
    \label{fig:omniglot_results}
\end{figure}

As an initial experiment, we provide more insight into the behaviour of methods designed to overcome catastrophic forgetting, motivating the use of online EWC. Figure \ref{fig:omniglot_forgetting} shows how the accuracy on the initial Omniglot alphabet varies over the course of training on the remaining 49 alphabets. The results allow for several interesting observations. Most importantly, we do not observe a significant difference between online EWC and EWC, despite the additional memory cost of the latter. The results for Learning Without Forgetting (LwF) show excellent results on up to 5 tasks, but the method struggles to retain performance on a large number of problems. The results for online EWC as applied within the P\&C framework are encouraging, yielding results comparable to the application of (online) EWC within a single neural network. As expected, the results for simple finetuning yield unsatisfactory results due to the lack of any protection against catastrophic forgetting.

\subsection{Assessing forward transfer}
\label{sec:forward_transfer}

\begin{figure}
\centering
\includegraphics[width=\linewidth]{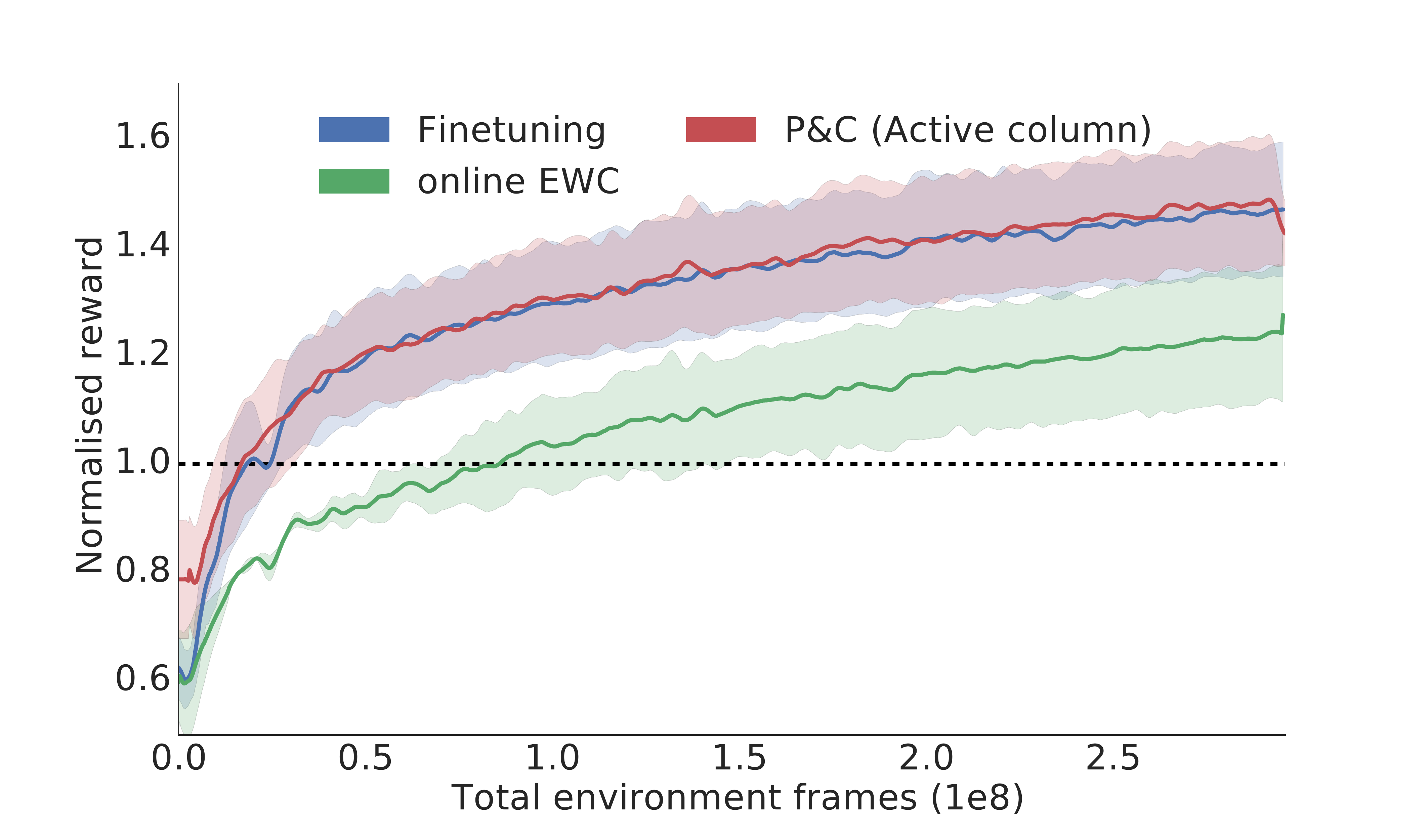}
\caption{Positive transfer on random mazes. Shown is the learning progress on the final task after sequential training. Results averaged over 4 different final mazes. All rewards are normalised by the performance a dedicated model achieves on each task when training from scratch. Best viewed in colour.}
\label{fig:lab_positive_transfer}
\end{figure}

Positive transfer in the context of transfer- and continual learning is typically understood as either an improvement in generalisation performance or more data-efficient learning. The latter is of great importance in problems where data acquisition can be costly, such as robotics \cite{rusu2016sim}. In order to assess the capability of P\&C to obtain positive transfer we show results for the navigation task in random mazes in Figure \ref{fig:lab_positive_transfer}. Specifically, we train on a held-out maze after having visited all 7 previous mazes. As the similarity between the tasks is high, we would expect significant positive transfer for a suitable method. Indeed, we observe both forms of transfer for all methods including online EWC (although to a lesser extent). P\&C performs on par with Finetuning, which in turn suffers from catastrophic forgetting. While online EWC does show positive transfer, the method underperforms when compared with Finetuning and P\&C.

\begin{table}
\small
\footnotesize
\caption{Positive Transfer on Atari. Shown is the relative performance after having trained on a various number of previous tasks.}
\centering
\begin{tabular}{lcccccc@{}}
    \toprule
    & \multicolumn{5}{c}{{\bf \% Single Task Performance}} \\
    \midrule
    Previous Tasks: & 1 & 2 & 3 & 4 & 5 \\
    \midrule
    P\&C (Active col, re-init)     & 127            & \textbf{129} & \textbf{125} & \textbf{129} & \textbf{128} \\
    P\&C (Active col)              & \textbf{131}   & 127          & 114          & 106          & 101 \\
    Finetuning                      & 117          & 125          & 117          & 105          & 98 \\
    EWC                             & 55           & 53           & 53           & 50           & 54 \\
    online EWC                      & 53           & 53           & 49           & 50           & 57 \\
    \bottomrule
\end{tabular}
\label{tab:positive_transfer_atari}
\end{table}

We show a summary of the same experiment on Atari in Table \ref{tab:positive_transfer_atari}. To quantify improved generalisation, we record the score after training on a unique task, having visited a varying number of different games beforehand. For P\&C, we report results obtained by the active column when parameters remain unchanged or are optionally re-initialised after a task has been visited (denoted re-init). 

In the case of a more diverse task set (Atari), both EWC versions show significant negative transfer, resulting in a decrease of final performance by over 40\% on average. While initially showing positive transfer, this effect vanished for Finetuning when more tasks are introduced. We observe the same effect for P\&C when parameters in the active column remain unchanged, suggesting only a small utilisation of connections to the knowledge base. 

Thus, as argued in Section \ref{sec:p_and_c}, we recommend re-initialising parameters in the active column, in which case P\&C continues to show significant positive transfer regardless of the number of tasks. Furthermore, the results show that positive transfer can indeed be achieved on Atari, opening the door for P\&C to outperform both online EWC and Finetuning when evaluating the overall performance of the method (see Section \ref{sec:overall_performance}).

In combination, these results suggest that the accumulation of Fisher regularisers indeed tends to over-constrain the network parameters. While this does not necessarily lead to negative transfer (provided high task similarity) we observe slower learning of new tasks compared to our method.

Conducting a similar experiment on Omniglot (see Figure \ref{fig:omniglot_forward_transfer}), we observed no generalisation improvement achieved by Progressive Nets or any other method across all alphabets when compared to training dedicated models per task. The effect of these methods is better described as ``avoiding negative transfer'', a phenomenon we continued to observe for EWC, online EWC \& Learning Without Forgetting (LwF). However, the application of P\&C did results in faster learning, a claim which we support with additional results in the Appendix. Together, these observations pose an interesting challenge for P\&C on Omniglot. Assuming a similar lack of positive transfer, the framework can only provide improvements if the knowledge preservation mechanisms can maintain more performance than a direct application of online EWC in a single network.

\subsection{Evaluating overall performance}
\label{sec:overall_performance}

Motivated by these results, we now investigate the overall performance for all methods. In case of P\&C we evaluate the model using the knowledge base (i.e. after distillation) and thus show how the method performs when several components are used in conjunction.

We first report the average test performance across all 50 Omniglot alphabets in Table \ref{tab:omniglot}, allowing for up to 5 re-visits of each alphabet (maintaining a fixed order during training). Importantly, we train until convergence on each visit. In order to provide a competitive comparison, we also include results achieved by less scalable methods. Progressive Nets (immune to forgetting) and the averaged results obtained by training a single model on each task (allowing no transfer) serve as such. All hyperparameters are optimised for maximum performance after five visits. We also show how the performance varies when 5 random task permutations are considered.

\begin{table*}[h]
\small
\footnotesize
\centering
\caption{Results on sequential Omniglot. Shown is the performance on all tasks after training. Results show mean and std. dev over task permutations.}
\begin{tabular}{lcccccr@{}}
    \toprule
    {\bf Model}  &
    \multicolumn{5}{c}{{\bf Test Accuracy}} & {\bf \#Parameters } \\
    Passes: & {\bf 1} & {\bf 2} & {\bf 3} & {\bf 4} & {\bf 5} & \\
    \midrule
    Single model per Task                    & 88.34                    & - & - & - & - & 5,680 K\\
    Progressive Nets                         & 86.50 {\tiny $\pm$  0.9} & - & - & - & - & 108,000 K\\
    \midrule
    Finetuning                               & 26.20 {\tiny $\pm$  4.6} & 42.40 {\tiny $\pm$  7.4} & 54.24 {\tiny $\pm$  7.1} & 60.84 {\tiny $\pm$  4.1} & 60.74 {\tiny $\pm$  3.8} & 217 K \\
    LwF ($\lambda=0.1$)                      & 62.06 {\tiny $\pm$  2.0} & 72.24 {\tiny $\pm$  2.6} & 68.44 {\tiny $\pm$  6.3} & 68.95 {\tiny $\pm$  3.0} & 66.48 {\tiny $\pm$  3.3} & 217 K \\
    EWC ($\lambda=12.5)$                     & 67.32 {\tiny $\pm$  4.7} & 71.92 {\tiny $\pm$  2.3} & 74.20 {\tiny $\pm$  2.8} & 74.46 {\tiny $\pm$  3.4} & 75.96 {\tiny $\pm$  3.2} & 11,100 K \\ 
    online EWC ($\lambda=17.5, \gamma=0.95)$ & 69.99 {\tiny $\pm$  3.2} & 73.46 {\tiny $\pm$  2.7} & 76.70 {\tiny $\pm$  1.9} & 79.26 {\tiny $\pm$  0.8} & 79.15 {\tiny $\pm$  1.9} & 446 K \\ 
    P\&C ($\lambda=15.0, \gamma=0.99)$       & 70.32 {\tiny $\pm$  3.3} & 76.28 {\tiny $\pm$  1.3} & 78.65 {\tiny $\pm$  1.4} & 80.13 {\tiny $\pm$  1.0} & 82.84 {\tiny $\pm$  1.4} & 659 K \\  
    \bottomrule
\end{tabular}
\label{tab:omniglot}
\end{table*}

\begin{figure*}
\centering
\includegraphics[width=\linewidth]{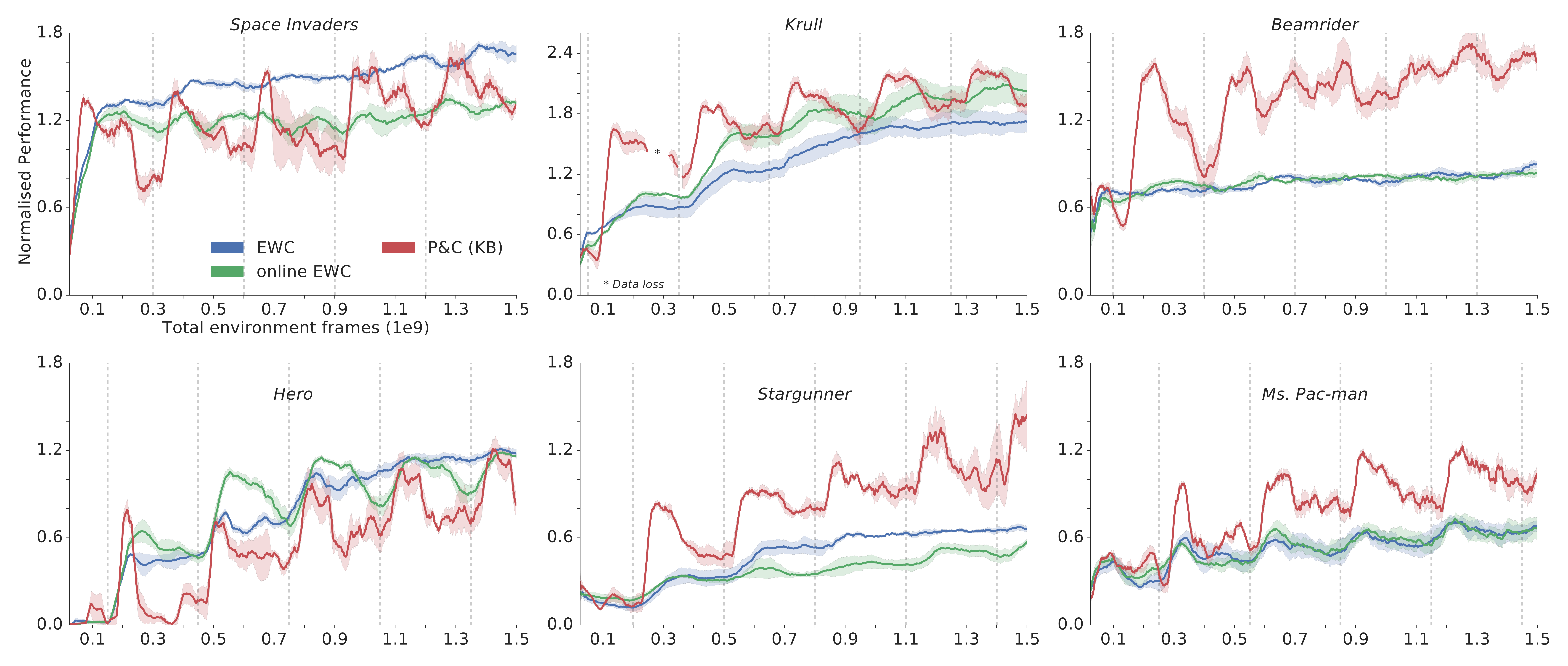}
\caption{Learning curves on Atari games. Each game is visited 5 times, allowing for training on 50m environment frames on each visit. Games are learned top to bottom left to right. Here KB: Knowledge base. Dashed vertical bars indicate re-visits to the task. Results averaged over random seeds. Best viewed in colour.}
\label{fig:atari_single_games}
\end{figure*}

As explained in Section \ref{sec:forward_transfer}, the performance of Progressive Nets is slightly lower than training a separate model per task. This is due to the observation shown in Figure \ref{fig:omniglot_forward_transfer}. Among the remaining methods, P\&C achieves the highest mean performance across all methods, although online EWC is competitive. The main observation explaining those results is a higher amount of negative transfer for online EWC, allowing some room for P\&C to take advantage of the two phases of learning. 

Another interesting observation is the difference in performance between EWC and the proposed online EWC, which we mainly observed when changing the amount of training on any given task for either method. We will discuss this in more detail below. LwF fails to achieve comparable results to either version of EWC which we attribute to the observations made in Figure \ref{fig:omniglot_forgetting}. 

Highlighting the lack of scalability of competing methods, we also include the number of parameters of each model in Table \ref{tab:omniglot}. Note that a large fraction of the parameters for Progressive Nets are due to the non-linear connections to each of the previous columns.

Moving onto experiments in Reinforcement Learning, we show learning curves for all Atari games in Figure \ref{fig:atari_single_games} after optimising all hyper-parameters for maximum final score across all games. In the case of P\&C, we only show rewards collected during the compress phase as the parameters remain unchanged when the active column is learning a new task. The results show a significant improvement for P\&C on several games while performing comparable (or slightly worse) on the remaining tasks.

Note that when choosing an appropriate regularisation term for the objective in \eqref{eq:ewc_loss} in the case of multiple visits to a task, allowing more forgetting to happen (i.e. choosing a lower $\gamma$) can lead to an overall higher performance. This is because a re-visit typically results in a higher score as the extent of EWC's capacity issues are weakened. This effect can be particularly well observed in the case of P\&C where an initial high amount of forgetting allows the knowledge base to perform overall better. Note that the regularisation strength $\lambda$ is not directly comparable between P\&C and both EWC variants as the scale of the loss (policy gradients or policy distillation) is different.

Thus we can conclude that P\&C is best used in domains that allow for some positive transfer in which it can show a large improvement over methods primarily designed to overcome catastrophic forgetting. In cases where this does not hold (e.g. Omniglot), P\&C can still show an improvement although online EWC on its own is a competitive method.

\section{Summary \& Discussion}
This work introduced Progress \& Compress, a framework designed to facilitate transfer in sequential problem solving while minimising the effects of catastrophic forgetting. The algorithm achieves a good trade-off between both objectives when combined with state-of-the-art-methods and works in a variety of challenging domains. 

Moreover, due to the generality of the proposed method, future methods to mitigate catastrophic forgetting should be easily integrable within our framework. Throughout this work, we made the assumption that the learner is aware of when changes in the task distribution occur, allowing for the computation of a new posterior approximation. However this is a relaxation of the more stringent requirement of knowing the identity of the current task that we hope can be exploited further to address the gradual drift problem described in Section \ref{sec:intro}. 

Additionally we use an online version of EWC very similar to the  proposal of \citet{huszar2017quadratic}. We add an explicit forgetting mechanism and provide empirical evidence suggesting that online EWC can perform well in practice.

\section*{Acknowledgements}
We would like to thank Jerome Connor, Nicolas Heess and Andrei A. Rusu for helpful discussions.

\nocite{shu2017}
\bibliography{p_and_c_rebuttal}
\bibliographystyle{icml2018}


\appendix

\newpage
\twocolumn[
\icmltitle{Progress \& Compress: A scalable framework for continual learning. Supplementary material}
\vskip 0.3in
]
\section{Retention of task performance for EWC and online EWC}
\begin{figure}
\centering
\includegraphics[width=\columnwidth]{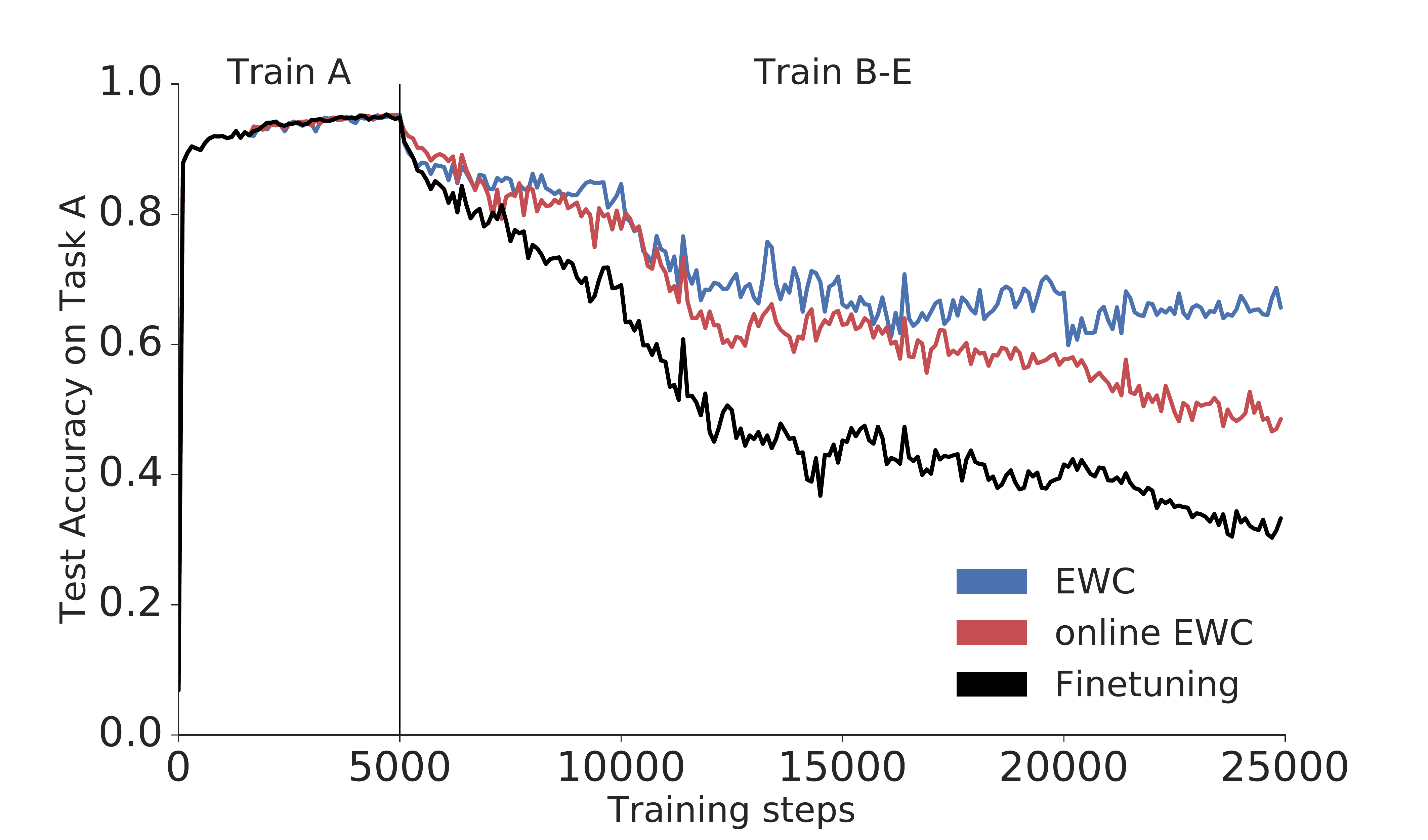}
\caption{Performance retention on permuted MNIST. Shown is the test accuracy 
         on an initial permutation (Task A) over the course of training on the
         remaining set of tasks (Tasks B-E).}
\label{fig:ewc_v_online_ewc_single_mnist}
\end{figure}

The difference between EWC and online-EWC is in their weighting of the past experiences, with EWC putting more weight on the initial tasks and online-EWC favouring the most recent past. For an optimal setting, where the optimisation converges and all penalty terms can be satisfied \cite{huszar2017quadratic}, online-EWC is often a better choice. However, it is likely that in difficult problems, the network (an agent) doesn't get enough time/training data to arrive at the optimal solution. 

We investigated this hypothesis in a series of experiments with a sequential learning of permuted-MNIST images, similar to the experiments shown in \cite{kirkpatrick2017overcoming}. In order to emulate learning difficult problems, we have not optimised the hyper-parameters, nor used any dropout or early stopping. Instead, we used a small MLP (layers consisting of 30-30-10 neurons, and Relu nonlinearities between the first two). 

Figure \ref{fig:ewc_v_online_ewc_single_mnist} demonstrates the retainment of the skill for the initial task (Task A) by EWC, online-EWC and pure SGD training (with no additional penalties), over the course of learning on a 
total of 5 permutations (Tasks A-E). As expected, EWC keeps higher accuracy for Task A.

\begin{figure}
\includegraphics[width=\columnwidth]{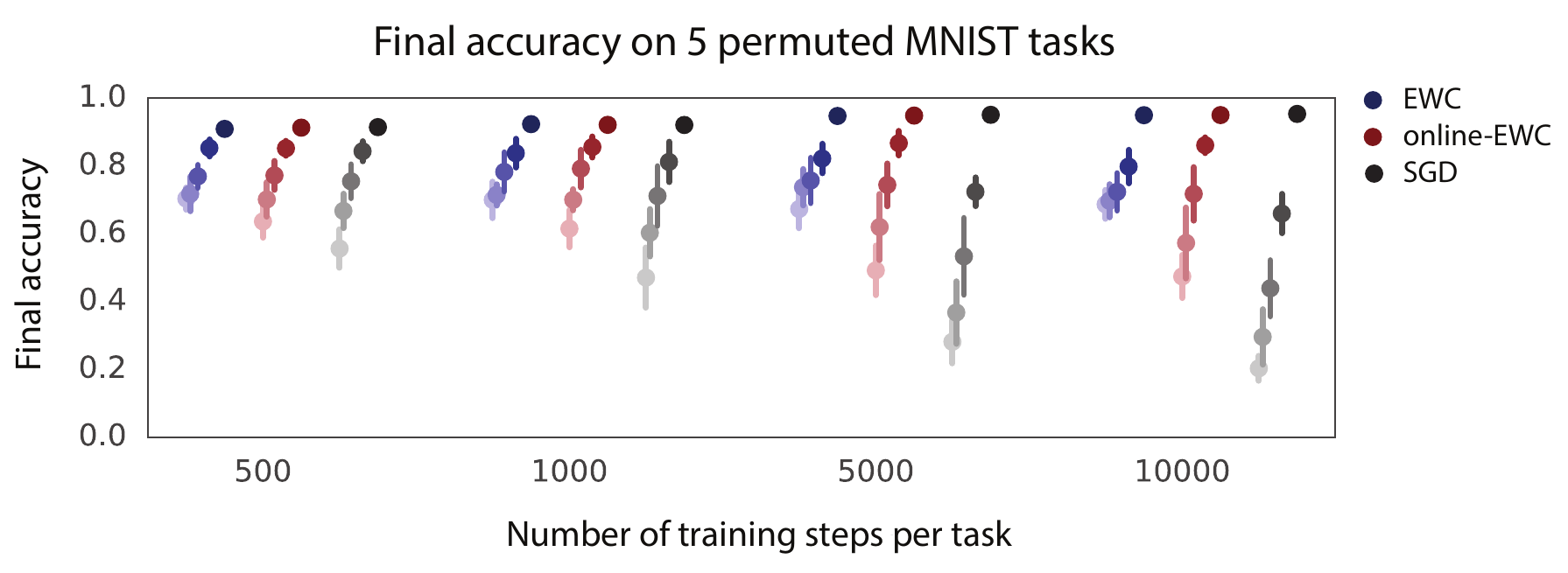}
\caption{Comparison of EWC, online-EWC and Finetuning}
\label{fig:ewc_v_online_ewc}
We ran the three training methods on 5 permuted-MNIST tasks (\cite{kirkpatrick2017overcoming}). The accuracy at the end of training is shown for each task with the fainter colours relating to the older tasks. The number of training steps on the x-axis relates to the number of minibatches of each task used for the training. In this regime (see text for details), EWC appears to be a better choice for a small number of training steps. 
\end{figure}

In Fig. \ref{fig:ewc_v_online_ewc}, we plot the final accuracy for each of the tasks (colour saturating from the faintest one representing Task A to the fully saturated for Task E), as a function of the number of training steps spent on each task. Here, we run 10 training sessions per fixed amount of training steps, generating new permutations for each training, but feeding exactly the same data to all methods (dot represents the mean and bars: 1 standard deviation (bar)). 

For a small number of training steps (500 and 1000, training over minibatches of size 32), the network benefits from holding on to the memories of the earlier tasks (the accuracy of EWC, i.e. all blue dots in the plot are higher than for the online EWC, the red dots). With more data (10,000 training steps), holding on to the initial parameters makes it more difficult to retain the most recent tasks (compare the dark blue dots of n=10,000 with n=500). In this example (with a relatively high learning rate $\eta=0.1$), the online EWC doesn't seem to find a good balance between the loss and penalties and the performance on older tasks is not well retained (faint red dots), although it's still better than using no penalty at all (grey dots).

\section{Faster learning on Omniglot}

While the experiments on Omniglot in the main text show that all considered methods fail to obtain a higher accuracy through sequential learning in the Omniglot case, improved data efficiency can indeed be observed for Progress\&Compress (P\&C).

\begin{figure}
    \centering
    \begin{subfigure}[b]{0.45\textwidth}
        \centering
        \includegraphics[width=\linewidth]{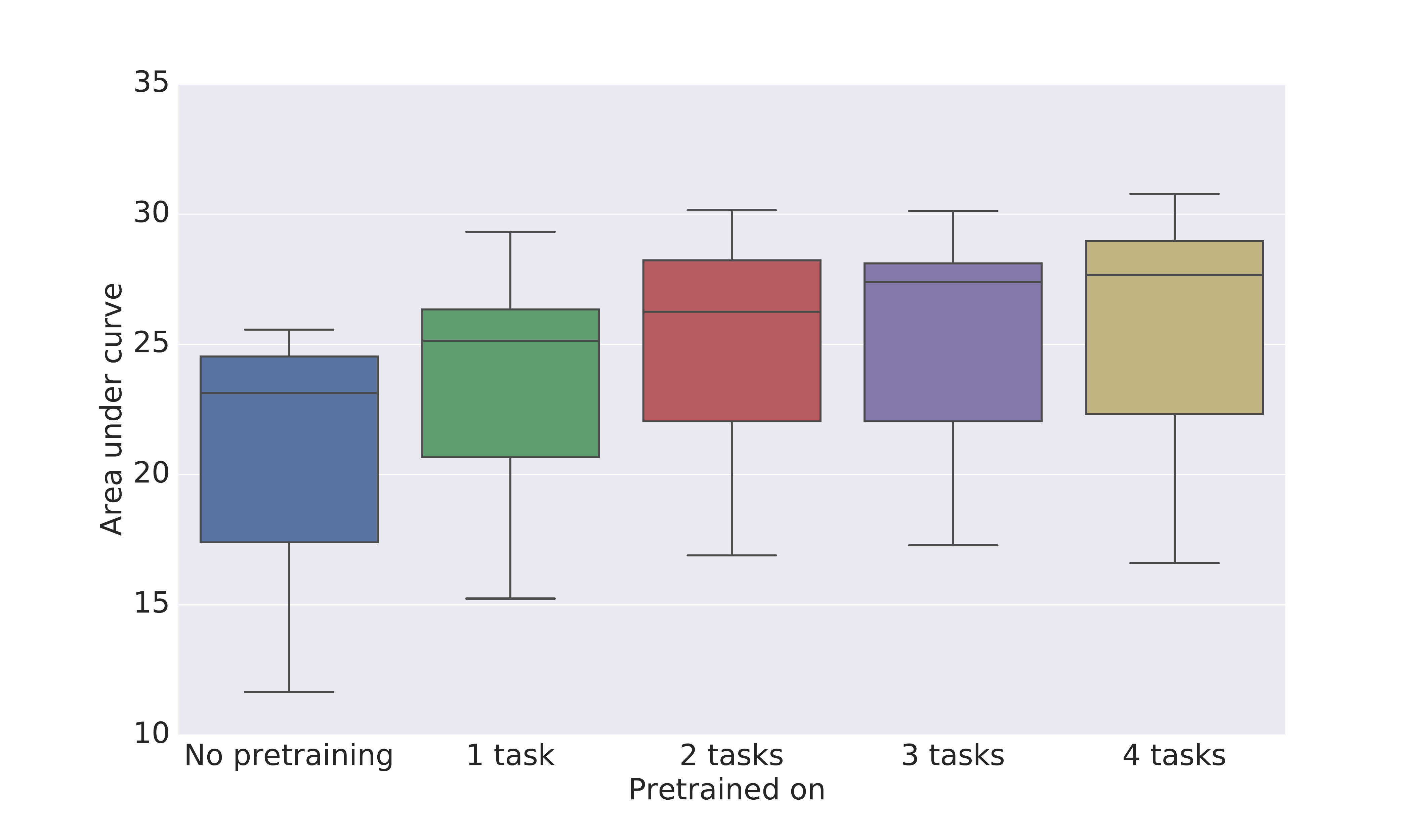}
        \caption{Area under the learning curve}
    \end{subfigure}
    \hfill
    \begin{subfigure}[b]{0.45\textwidth}
        \centering
        \includegraphics[width=\linewidth]{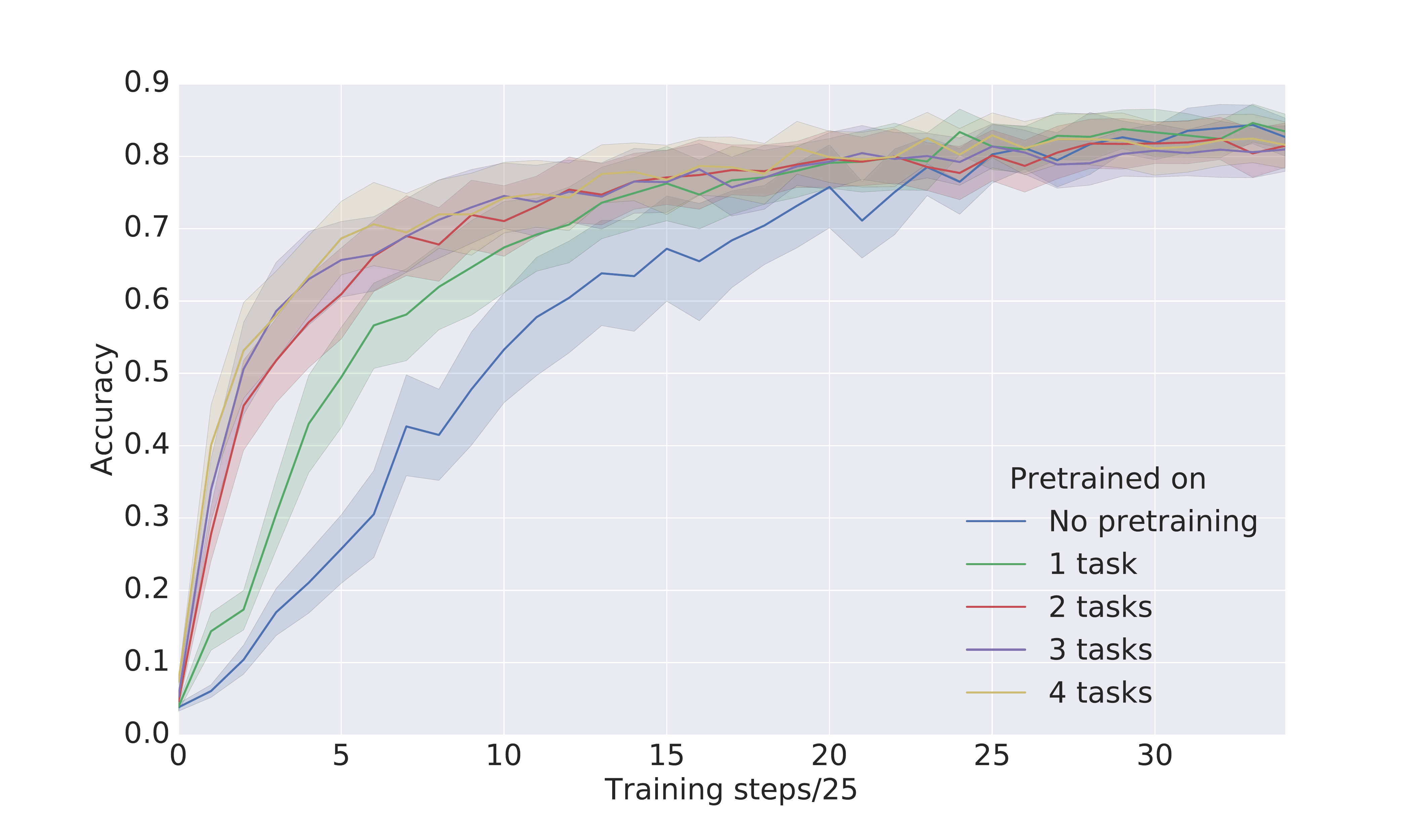}
        \caption{Averaged learning curves}
    \end{subfigure}
    \caption{Faster learning on Omniglot with Progress \& Compress. Results averaged over 10 alphabets.}
    \label{fig:omniglot_faster_learning}
\end{figure}

In order to test this in isolation, we trained P\&C (with online-EWC) on 10 unique Omniglot alphabets, after having learned up to 4 different tasks. We show both the averaged learning curves as well as the area under those learning curves in Figure \ref{fig:omniglot_faster_learning}. The results suggest that pre-training on a small number of tasks can greatly improve data efficiency, with this effect plateauing for more than 4 tasks.

\section{Experiment details}

\subsection{Omniglot}
In the Omniglot experiments, we used a convolutional network similar to the one introduced by \citet{vinyals2016matching}, ensuring each method has sufficient capacity to learn all tasks. Namely, the network consists of 4 blocks of $3\times3$ convolutions with 64 filters followed by a ReLU nonlinearity,
and $2\times2$ max-pooling before predicting class probabilities. In the case of P\&C and Progressive Nets, all network columns follow this architecture. As suggested in \cite{rusu2016progressive}, non-linear adapters for convolutional networks are implemented by replacing each linear layer by a $1\times1$ convolutions using an identical number of filters.

Similar to the setup proposed in \cite{koch2015siamese} we used a 60/20/20\% split to obtain train-/valid- and test-sets. In addition, we rescaled all images to $28\times28$ and augment the dataset by including 20 random permutations (rotations and shifting) for each image. Note that since we are not treating Omniglot in the usual few-shot learning fashion, we do not distinguish between train and test alphabets. 

For all considered models, we used a batch size of 32 and perform 2500 training updates with Stochastic Gradient Descent and a fixed learning rate of 0.1 (0.05 during distillation), which we found sufficient to learn each alphabet separately from scratch.

For EWC, online EWC and P\&C, we chose the regularisation strength $\lambda$ and forgetting coefficient $\gamma$ by running a grid search for $\lambda$=[10.0, 12.5, 15.0, 17.5, 20.0, 22.5, 25.0] and $\gamma$=[0.7, 0.8, 0.9, 0.95, 0.99]. For Learning Without Forgetting (LwF) we tried $\lambda$=[0.05, 0.1, 0.15, 0.2, 0.25, 0.3]. For distillation within P\&C and LwF, we found a softmax temperature $\tau=2.0$ to work best. All hyperparmeters were tuned by maximising the averaged performance over all tasks using aforementioned validation set.

Note that we use the same network and optimisation settings throughout all experiments. This is with the exception of results showing positive transfer and forgetting in isolation, in which case we fix $\lambda$ for all EWC methods to provide a fair comparison.

\subsection{Atari \& Navigation tasks}

For both Atari\&Navigation tasks we use the same network as in \cite{mnih2013playing}, adopting it to actor-critic algorithms by estimating both value and policy through linear layers connected to the final output of a shared network. During optimisation, we use a batchsize of 20, unroll length of 20 and update the model parameters with RMSProp  (using $\epsilon=0.1$), linearly annealing the learning rate to 0 over the course of training. For navigation mazes, we used an initial learning rate of $\alpha=0.004$ and entropy cost $\beta=0.003$. For Atari games, we used $\alpha=0.0006$ and $\beta=0.01$. In both cases, we receive RGB environment frames as $84\times84\times3$ tensors. As is common, we apply each action 4 times to the environment.

Furthermore, we use clip rewards so that the maximum absolute reward is 1.0. We also use a baseline cost of 0.5 in the policy gradient loss. The discounting factor was set to 0.99.

EWC was separately tuned choosing $\lambda$ from [500, 1000, 1500, 2000, 2500, 3000]. As the scale of the losses differ, we selected $\lambda$ for online EWC as applied in P\&C among [25, 75, 125, 175]. We use 100 minibatches of equal size to estimate the diagonal Fisher.

\end{document}